# Attention is Not *Always* What You Need: Towards Efficient Classification of Domain-Specific Text

## Case-study: IT Support Tickets

Yasmen Wahba [1](✉), Nazim Madhavji[1], and John Steinbacher [2]

[1] Western University, London ON, Canada
ywahba2@uwo.ca, nmadhavji@uwo.ca
[2] IBM Canada, Toronto, ON, Canada
jstein@ca.ibm.com

**Abstract.** For large-scale IT corpora with hundreds of classes organized in a hierarchy, the task of accurate classification of classes at the higher level in the hierarchies is crucial to avoid errors propagating to the lower levels. In the business world, an efficient and explainable ML model is preferred over an expensive black-box model, especially if the performance increase is marginal. A current trend in the Natural Language Processing (NLP) community is towards employing huge pre-trained language models (PLMs) or what is known as self-attention models (e.g., BERT) for almost any kind of NLP task (e.g., question-answering, sentiment analysis, text classification). Despite the widespread use of PLMs and the impressive performance in a broad range of NLP tasks, there is a lack of a clear and well-justified need to as why these models are being employed for domain-specific text classification (TC) tasks, given the monosemic nature of specialized words (i.e., jargon) found in domain-specific text which renders the purpose of contextualized embeddings (e.g., PLMs) futile. In this paper, we compare the accuracies of some state-of-the-art (SOTA) models reported in the literature against a Linear SVM classifier and TFIDF vectorization model on three TC datasets. Results show a comparable performance for the LinearSVM. The findings of this study show that for domain-specific TC tasks, a linear model can provide a comparable, cheap, reproducible, and interpretable alternative to attention-based models.

## 1 Background

As the volume of information available on the Internet increases, there is a growing interest in developing tools to rapidly find, filter, and better manage these electronic resources. TC which is the task of classifying text (e.g., tweets, news, and customer reviews) into different categories (i.e., tags), is a crucial component in many information organization and management tasks. With the ubiquity of data, the need to fully automate text classification methods becomes vital. In IT service management, TC can be applied for many purposes. One of these is classifying IT support tickets. A support ticket describes an issue faced by the customer that is submitted as a bug report to the IT support team. Support agents spend a significant amount of time manually classifying incoming tickets and there is no reference to best practices based on historical data.

A current trend in the NLP community is towards employing deep learning (DL) based models or PLMs for several NLP tasks including text classification. This trend is stimulated by the prevalence of 'Leaderboards'. A leaderboard is the main component of machine learning competitions that are hosted by large companies such as Netflix or popular online platforms such as Kaggle [1]. The 'Leaderboard' ranks the best submissions for the participating teams by their accuracy scores (i.e., classifier's performance). Recently, NLP leaderboards are dominated by PLMs which achieve SOTA results on several benchmarks such as GLUE [2] or individual datasets such as SQuAD [3].

Another reason why attention-based models (e.g., BERT) are favorable in the NLP community is that they have an enormous number of trainable parameters which enables these models to encode a substantial amount of linguistic knowledge/structure. This encoded linguistics knowledge is beneficial to address the issue of word polysemy [4]. Polysemy is a phenomenon that is getting much attention in the literature recently, where a word could have different meanings and thus should have different vector embeddings and not just one. However, we argue that in a domain-specific text that contains a large number of technical words (i.e., jargon), a word has a more precise meaning (i.e., monosemy) [5].

Despite the impressive success of PLMs in a broad range of NLP tasks, there is a lack of a clear and well-justified need to as why these models are being employed for domain-specific TC tasks [6][7][8] given the following:

• Most text classification problems are linearly separable [9][10]. This is because text datasets are characterized by a high number of features that inaugurate the linear separability of the data.

• The large gap between the pre-training cloze-style formulation and objectives (e.g., predict target words) and the downstream objectives (e.g., classification) limit the ability to fully utilize the knowledge encoded in PLMs [11].

• The level of polysemy in domain-specific (i.e., specialized) text is low [12] which defeats the purpose of contextualized embeddings that aim to capture word polysemy and provide more than one embedding for a single word.

• Domain-specific terms are challenging for PLMs since there are few statistical clues in the underlying training corpora [13][14].

Our work with IT support agents for a large industrial IT partner to classify customer support tickets has shed light on two main real-world concerns these large corporations face with DL-based models. The first concern is *reproducibility* which creates trust and credibility with the ML model. A recent literature survey [15] reveals that the reproducibility of DL models remains a major concern. Due to the randomness of the hyperparameters and weights used in the training stage for DL models and non-determinism in the hardware (i.e., computing resources like GPUs), it is challenging to reproduce these models [15][16].

The second concern is the *interpretability* of the results. While the field of eXplainable Artificial Intelligence (XAI) has regained the attention of researchers over the past few years [17][18][19], the explanations they provide are not accurate (i.e., low fidelity) [20]. Cynthia Rudin [20] argues that if the explanations were completely faithful to what the original model computes, we would not need the original model in the first place and the explanations should suffice.

Thus, if for a certain task, there exists a linear interpretable model that offers the same accuracy as a black-box model with less computational power and faster training times, then it is preferable to use the simple linear model. We note that the work here is confined to the specific task of classifying domain-specific text and not other NLP tasks. For instance, sentiment classification (or sentiment analysis) is one use-case of text classification; however, words that express sentiment have fuzzy meanings (i.e., polysemy). An example of how the word 'funny' could be classified as 'happy' or 'suspicious' is found in [21].

## 2    Monosemy (i.e., Absence of ambiguity)

The term 'Monosemy' from the Greek roots: mono ("one") and semainein ("to signify") -- stands for words with only one meaning [12]. It is the opposite of polysemy where words could have more than one meaning [22].

For domain-specific text, the monosemic nature of words is intrinsically linked to the technical/specialized vocabulary (e.g., DNS). The reason behind this is that scientific terms need a precise meaning in order to function and be easily recognized [12].

The following table (Table 1) shows a sample of pre-processed tickets (see the first column) from our private support tickets dataset. The second column highlights a specialized (i.e., domain-specific) word that could have different possible meanings if appeared in a different context. However, the actual meaning (in the third column) is the only logical/intended meaning for the word in the context of a ticketing system.

**Table 1**. The monosemic nature of some words that appear in the IT support tickets dataset, their actual meaning in the text, and another possible meaning.

| Examples of support tickets | Specialized word | Actual meaning in the text | Other possible meaning |
|---|---|---|---|
| Subscription account link **cloud** … | Cloud | A system hosting software services | A visible mass of particles of condensed vapor |
| Cancel line item **whiskey** | Whiskey | A user-interface | A drink |

| **Slave** node serve customer traffic … | Slave | A device | A person held in forced servitude |
| --- | --- | --- | --- |
| **Host** freeze case brings production back … | Host | A computer | A person who talks to guests on a program |
| Good regard organization **space** resource field … | Space | A container | The region beyond the earth's atmosphere |
| Clear **cookie** success | Cookie | A file | A cake |
| Make **soap** connection web team … | Soap | Simple Object Access Protocol | A cleansing agent |
| **Boot** access web service … | Boot | Verb- to reload | A footwear |

## 3 Results

Our experiments were evaluated on the following three datasets:

1. 20NewsGroup [23]: a public dataset consisting of 18,846 documents, categorized into 20 groups. We note that some research paper uses a version of this dataset with only four major categories, hence their results were not included in this paper.
2. BBC News [24]: a public dataset originating from BBC News. It consists of 2,225 documents, categorized into 5 groups, namely: business, entertainment, politics, sport, and tech.
3. IT support tickets: a private dataset obtained from a large IT industrial partner. It is composed of real customer issues related to a cloud-based system. It consists of 194,488 documents categorized into 12 classes.

**Table 2.** Accuracy results of SOTA models reported in the literature on two TC datasets against a Linear SVM with the highest accuracies in bold.

| Dataset | Model | Accuracy | Reference |
| --- | --- | --- | --- |
| 20NewsGroup (20 classes) | TFIDF with Naive-Bayes | 81.69 | [25] |
| | GloVe+Average | 80.43 | [25] |
| | GloVe+Attention | 81.65 | [25] |
| | LSTM+CNN | 79.74 | [25] |
| | BiLSTM+Max | 83.02 | [25] |
| | BiLSTM+Attention | 81.76 | [25] |
| | Universal Sentence Encoder (USE) | 81.76 | [25] |
| | ULMFiT | 82.4 | [25] |
| | Hierarchical Attention Network (HAN) | 85.01 | [25] |
| | BERT | 85.78 | [25] |
| | DistilBERT | 85.43 | [25] |
| | fastText | 79.4 | [26] |
| | MS-CNN | 86.1 | [27] |
| | Text GCN | 86.3 | [28] |
| | TensorGCN | 87.74 | [29] |
| | Simplified GCN | 88.50 | [30] |
| | MLP over BERT | 85.5 | [27] |
| | LSTM over BERT | 84.7 | [27] |
| | LEAM | 81.91 | [31] |
| | CogLTX (Glove init) | 87.0 | [32] |
| | BoW + SVM | 63.0 | [32] |
| | Bi-LSTM | 73.2 | [32] |
| | **RoBERTaGCN** | **89.5** | [33] |
| | **SVM+TFIDF** | **90.0** | |
| BBC News (5 classes) | *BERT* | 97 | [34] |
| | *DistilBERT* | 97 | [34] |
| | *XLM* | 97 | [34] |
| | *RoBERTa* | **99** | [34] |
| | *XLNET* | 98 | [34] |
| | TFIDF with Naive-Bayes | 95.73 | [24] |
| | GloVe+Average | 94.16 | [25] |
| | GloVe+Attention | 95.28 | [25] |
| | LSTM+CNN | 96.18 | [25] |
| | BiLSTM+Max | 95.73 | [25] |
| | BiLSTM+Attention | 96.63 | [25] |
| | Universal Sentence Encoder (USE) | 96.63 | [25] |
| | ULMFiT | 97.07 | [25] |
| | Hierarchical Attention Network (HAN) | 97.75 | [25] |
| | **BERT** | **98.2** | [25] |
| | DistilBERT | 97.3 | [25] |
| | **SVM+TFIDF** | **98.0** | |
| IT support tickets (12 classes) | BERT | 0.79 | |
| | DistilBERT | 0.78 | |
| | XLM | 0.79 | |
| | RoBERTa | 0.79 | |
| | **SVM+TFIDF** | **0.79** | |

Table 2 shows that the linear SVM [35] is comparable to SOTA models reported in the literature. For our private dataset of IT support tickets, we performed the basic pre-processing (i.e., cleaning) steps for our tickets, and for the PLMs we fine-tuned the PLMs for 3 epochs as we suffered from overfitting when the number of epochs exceeded 3. Also, for the vectorization step, TFIDF was used with the n-gram setting set to 3 (i.e., tri-grams).

It is to be noted that for the 20NewsGroup dataset, accuracies higher than 90% are for authors using only the four major categories (comp, politics, rec, and religion) out of the original 20 categories. For instance, [36] reported an accuracy of 96.5% using a 2D Convolutional Filter. Similarly, [37] reported an accuracy of 96.49% using recurrent convolutional neural networks. However, we note that they use only four major categories (comp, politics, rec, and religion) out of the original 20 categories for the 20NewsGroup. Hence, we strongly recommend renaming this dataset to include the number of categories (e.g., 20NewsGroup-4) to denote using only four categories and to provide a fair comparison.

## 4   Conclusion

Domain-specific TC is a fundamental problem in NLP. However, the current trend is towards using computationally expensive models such as PLMs to solve any NLP task. This paper argues the need for employing such huge models in industrial settings where reproducibility and interpretability are important factors for successful deployment. We note that the work here does not underestimate the power of DL models for classification tasks. It is simply arguing against the use of such complex models for classifying specialized text where words have precise meanings. The study indicates a comparable performance for a traditional linear model such as SVM and fine-tuned SOTA models reported in the literature for TC tasks and especially domain-specific datasets. We encourage the replication of this study on more domain-specific datasets for greater validity of the findings.